\documentclass[10pt,twocolumn,letterpaper]{article}

\usepackage{cvpr}              % To produce the CAMERA-READY version
\usepackage{multirow}
\definecolor{cvprblue}{rgb}{0.21,0.49,0.74}
\usepackage[pagebackref,breaklinks,colorlinks,allcolors=cvprblue]{hyperref}

\title{DreamDance: Animating Human Images \\ by Enriching 3D Geometry Cues from 2D Poses}

\author{%
  Yatian Pang\textsuperscript{1,3},
  Bin Zhu\textsuperscript{1},
  Bin Lin\textsuperscript{1},
  Mingzhe Zheng\textsuperscript{4},
  \\
  Francis E. H. Tay\textsuperscript{3},
  Ser-Nam Lim\textsuperscript{5,6},
  Harry Yang\textsuperscript{4,6},
  Li Yuan\textsuperscript{1,2,†}
  \\
  \\
  \textsuperscript{1} Peking University \quad
  \textsuperscript{2} PengCheng Laboratory \quad
  \textsuperscript{3} NUS \quad
  \textsuperscript{4} HKUST \quad
  \textsuperscript{5} UCF \quad
  \textsuperscript{6} Everlyn AI \quad
}  

\begin{document}
\maketitle

\begin{abstract}
In this work, we present \textbf{DreamDance}, a novel method for animating human images using only skeleton pose sequences as conditional inputs. Existing approaches struggle with generating coherent, high-quality content in an efficient and user-friendly manner. Concretely, baseline methods relying on only 2D pose guidance lack the cues of 3D information, leading to suboptimal results, while methods using 3D representation as guidance achieve higher quality but involve a cumbersome and time-intensive process. To address these limitations, DreamDance enriches 3D geometry cues from 2D poses by introducing an efficient diffusion model, enabling high-quality human image animation with various guidance. Our key insight is that human images naturally exhibit multiple levels of correlation, progressing from coarse skeleton poses to fine-grained geometry cues, and further from these geometry cues to explicit appearance details. Capturing such correlations could enrich the guidance signals, facilitating intra-frame coherency and inter-frame consistency. Specifically, we construct the TikTok-Dance5K dataset, comprising 5K high-quality dance videos with detailed frame annotations, including human pose, depth, and normal maps. Next, we introduce a Mutually Aligned Geometry Diffusion Model to generate fine-grained depth and normal maps for enriched guidance. Finally, a Cross-domain Controller incorporates multi-level guidance to animate human images effectively with a video diffusion model. Extensive experiments demonstrate that our method achieves state-of-the-art performance in animating human images. Webpage: \url{https://pang-yatian.github.io/Dreamdance-webpage/}
\vspace{-1.2em}
\end{abstract}    
\section{Introduction}
\label{sec:intro}

\begin{figure}[ht]
    \centering
    \includegraphics[width= \linewidth]{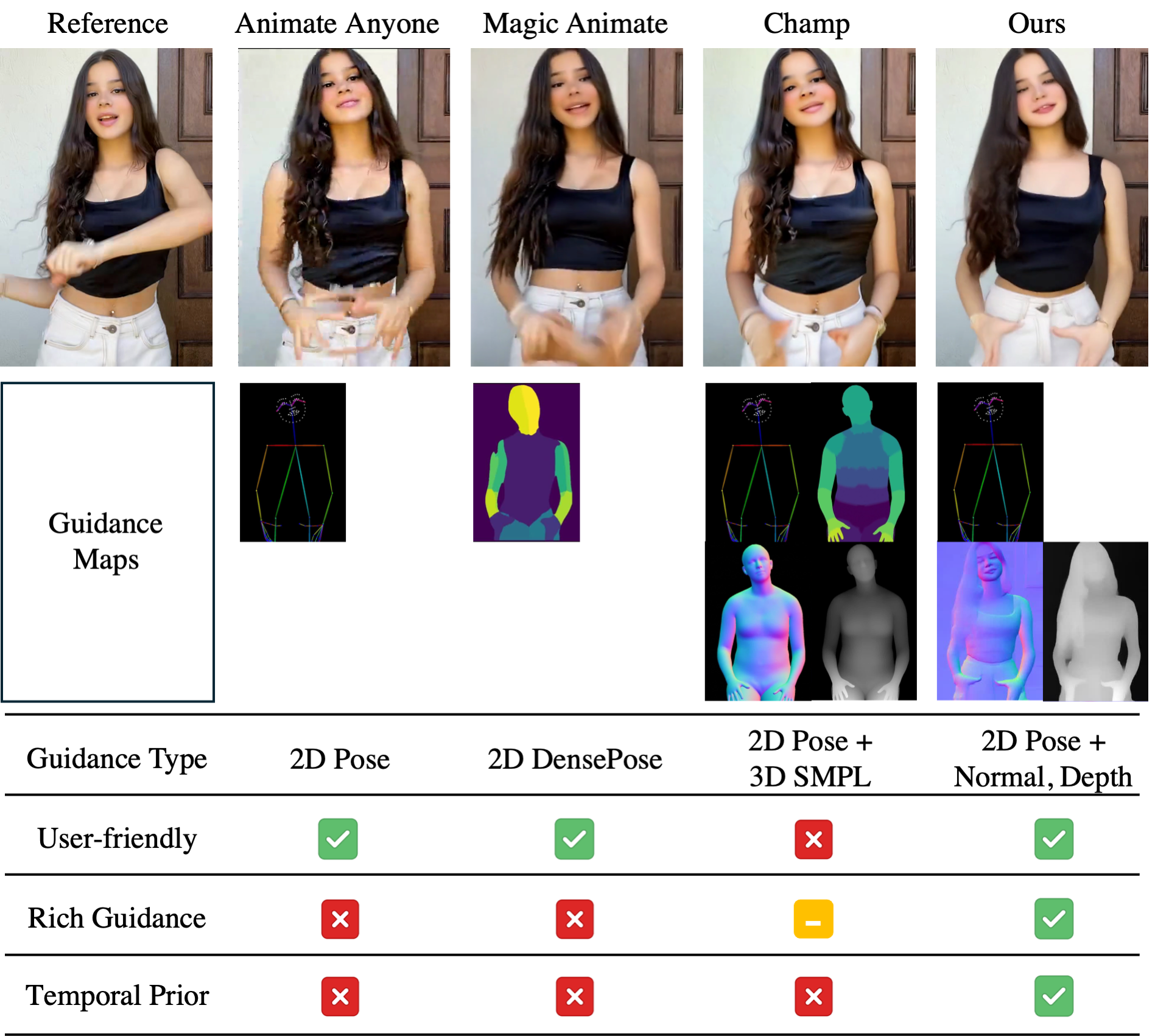}
    \vspace{-2em}
    \caption{A summarized comparison between DreamDance and the baseline methods.}
    \vspace{-2em}
    \label{fig:head}
\end{figure}

Human image animation refers to generating dynamic and realistic videos from static human images following a sequence of motion control signals. This field has attracted significant academic interest due to its diverse applications across industries, including film production, social media, and online retail. Despite the rapid development of generative artificial intelligence, human image animation remains challenging because it requires a comprehensive understanding of both intra-frame coherency and inter-frame consistency.

Recently, diffusion models~\cite{ho2020denoising,song2020denoising} have demonstrated remarkable potential in image~\cite{rombach2022high,brooks2022instructpix2pix,ruiz2022dreambooth,Peebles2022DiT} and video~\cite{blattmann2023stable,blattmann2023align,yang2024cogvideox,zhou2022magicvideo,zeng2024make} generation, unlocking new possibilities for human image animation tasks. Pioneer works such as Disco~\cite{wang2023disco} and Dreampose~\cite{karras2023dreampose} leverage input images and pose conditions to generate target frames sequentially. However, due to the lack of temporal learning capabilities, these methods often suffer from flickering artifacts and temporal inconsistency across frames. To address these limitations, more recent methods such as MagicAnimate~\cite{xu2024magicanimate} and AnimateAnyone~\cite{hu2024animate} incorporate temporal attention blocks into the diffusion network to enhance temporal coherence. Despite this improvement, their generation quality remains constrained by coarse control conditions, leading to inconsistent and incoherent visual outputs.
To further improve control precision, Champ~\cite{zhu2024champ} introduces a 3D parametric human model SMPL~\cite{loper2023smpl} to provide additional guidance. By first generating SMPL motions and then rendering them into normal and depth maps, this approach integrates geometric information with pose sequences for human image animation. However, SMPL-based methods present significant drawbacks. First, generating SMPL motions is cumbersome and not user-friendly, as it typically relies on predictions from existing videos, which limits editability and flexibility compared to the more intuitive manipulation of poses using open-source tools like the GUI~\cite{sd-webui-3d-open-pose-editor}. Second, the independence between SMPL and pose models can lead to misalignment between control signals, resulting in imprecise adherence to the intended pose sequences. Additionally, SMPL's rendering primarily emphasizes body geometry while neglecting crucial visual details, such as clothing and hair, which can contribute to temporal incoherence and visual inconsistencies. Furthermore, these methods take image diffusion models as base models, which lack a strong temporal prior, resulting in poor temporal consistency in the generated videos. We summarize existing methods in Figure~\ref{fig:head}. 

To this end, we propose \textbf{DreamDance}, a unified framework for animating human images using only skeleton pose sequences as original guidance signals. Our key insight is the recognition of multi-level correlations inherent in human images, extending from coarse skeleton poses to fine-grained geometry cues, and further to explicit appearance details. Capturing these correlations could enrich the guidance signals, thereby enhancing both intra-frame coherence and inter-frame consistency in the animation process.
To achieve this, we first construct a dataset, TikTok-Dance5K, comprising 5,000 high-quality dance videos, each annotated with human pose, depth, and normal maps for every frame.  Building on this dataset, we propose a novel framework comprising two diffusion models. Specifically, the Mutually Aligned Geometry Diffusion Model generates detailed depth and normal maps to enrich the guidance signals. To ensure the robustness and effectiveness of the generated guidance signals, we introduce geometry attention and temporal attention modules, which align guidance across modality and temporal dimensions. With the enriched geometry cues, the Cross-Domain Controlled Video Diffusion Model utilizes a cross-domain controller to integrate multiple levels of guidance, animating human images with high quality.  Besides, we implement a robust conditioning scheme to reduce the impact of error accumulation in our two-stage generation pipeline.
Extensive experiments demonstrate that DreamDance achieves state-of-the-art performance in animating human images, significantly improving both visual coherence and temporal consistency compared to existing methods. We again highlight the comparison in Figure~\ref{fig:head} and show that by eliminating the need for complex and rigid 3D models, our approach provides greater flexibility, ease of use, and finer control over the animation process, establishing it as a highly effective solution for animating human images.

To sum up, this work contributes in the following ways: 

\begin{itemize} 

\item We construct the TikTok-Dance5K dataset, comprising 5,000 high-quality human dance videos with comprehensive visual annotations. It will be made publicly available to support further research in the field.

\item We propose the Mutually Aligned Geometry Diffusion model, which generates detailed depth and normal maps to enrich guidance signals. To ensure robustness, we incorporate geometry and temporal attention modules that align geometric information across modality and temporal dimensions.

\item We introduce the Cross-Domain Controlled Video Diffusion Model, which utilizes a cross-domain controller to integrate multiple levels of guidance for high-quality human image animating. A robust conditioning scheme further mitigates the effects of error accumulation in our two-stage generation pipeline.

\item Our extensive experiments demonstrate that our framework achieves state-of-the-art performance in animating human images, setting a new benchmark for visual coherence and temporal consistency compared to existing methods. 

\end{itemize}
\section{Related Works}
\label{sec:related_works}
\subsection{Diffusion Models for Image Generation}

Diffusion models have achieved significant success in generating images given various conditions. Notably, Stable Diffusion~\cite{rombach2022high} has demonstrated its potential in producing high-quality images from text prompts as the conditioning input. To introduce additional control over the generation process, ControlNet~\cite{zhang2023adding} proposes training an additional copy of the pre-trained diffusion model to incorporate extra control signals, such as pose, depth, and canny edges. Subsequently, many works~\cite{ye2023ip,mou2024t2i,shi2024instadrag} propose incorporating more complex control signals for enhanced controllability in image generation. In this work, we focus on using pose as the control signal and propose generating fine-grained depth and normal maps to provide supplementary guidance. Several works~\cite{long2023wonder3d,pang2024envision3d,qiu2024richdreamer} propose generating geometry maps with diffusion models to improve the quality of the generated 3D geometry. Other works~\cite{fu2024geowizard,ke2023repurposing} propose to utilize diffusion models for depth and normal estimation tasks. To generate precise human images, Hyperhuman~\cite{liu2023hyperhuman} proposes a latent structural diffusion model to jointly capture the human image appearance and geometry relationship. Their diffusion model consists of different expert branches and only supports text as input conditions. In our work, we propose a novel geometry diffusion model by introducing a geometry attention mechanism and utilizing a reference net for fine-grained reference image control. We also incorporate temporal attention to enable the geometric generation along the temporal dimension.  

\subsection{Diffusion Models for Video Generation}
The great potential of diffusion models in image generation has inspired extensive research in applying them to video generation and other domains.~\cite{wu2023tune,yuan2024magictime,khachatryan2023text2video,shi2024bivdiff,yu2023animatezero,geyer2023tokenflow,chen2024videocrafter2,wang2024leo,esser2023structure,qi2023fatezero,zhang2024show,huang2024vbench,guo2024i2v,peng2024controlnext,tang2024cycle3d,jin2023act,jin2025local,pang2024envision3d,chen2024multi,zongying2024taxdiff,zhang2025repaint123}. Pioneer works~\cite{esser2023structure,hong2022cogvideo,ho2022video,khachatryan2023text2video,singer2022make} focus on extending image-based diffusion models using temporal modules to address the dynamics of video sequences and generating videos. Video LDM~\cite{blattmann2023align} first pre-train the model on images and then fine-tune it on video data by incorporating temporal layers. Animatediff~\cite{guo2023animatediff} add additional motion modules into pre-trained text-to-image diffusion models and train them with video data. Stable Video Diffusion (SVD)~\cite{blattmann2023stable} is a latent video diffusion model for high-resolution image-to-video generation.  The authors identify and evaluate three different stages for successful training of the video diffusion model and explore a unified strategy for curating video data.  In this work, we construct SVD-ControlNet and take Stable Video Diffusion as the base model because it possesses a strong temporal prior and temporal consistency compared to training from scratch or adapting from image diffusion models.
% However, the generation quality is limited by coarse control conditions, presenting incoherent content. 

\subsection{Diffusion Models for Human Image Animation}

Human image animation aims to generate dynamic and realistic human motion videos given a static human image and a sequence of control signals~\cite{albahar2023humansgd,shao2024human4dit,cao2024dreamavatar,chan2019dance,ren2020deep,yu2023bidirectionally,mimicmotion2024,wang2024vividpose,wang2024unianimate,geng2023affective,wang2023agentavatar,wang2024instructavatar,ma2024follow,guo2024liveportrait,xie2024x,jiang2024mobileportrait,xue2024follow,feng2023dreamoving,tu2024motionfollower,kim2024tcan,zhai2024idol,xu2024tunnel,fang2024vivid,he2024wildvidfit,huang2024make,wang2024videocomposer,men2024mimo,tian2024emo,he2024id,xu2024ootdiffusion,wang2024humanvid}. Recent works utilize the successful diffusion models for this task. DreamPose~\cite{karras2023dreampose} and DISCO~\cite{wang2023disco} modify the diffusion model to integrate the information from the reference image and control signal. However, they animate the human image in a frame-by-frame manner, which lacks temporal consistency. The following works AnimateAnyone~\cite{hu2024animate}, MagicAnimate~\cite{xu2024magicanimate}, and MagicPose~\cite{chang2023magicdance} propose a similar framework that uses a reference network for human feature injection and also utilizes motion modules to enhance temporal consistency. The recent work Champ~\cite{zhu2024champ} introduces the use of a 3D parametric human model, SMPL, to provide additional conditions for human image animation by generating normal and depth maps from SMPL motions. However, this approach has limitations, including the complexity of creating SMPL motions, potential misalignment between SMPL and pose models, and the omission of important details like outfits and hair, leading to less coherent generated content. Our method addresses these limitations by proposing a Mutually Aligned Geometry Diffusion Model to jointly capture human image appearance and geometry relationship, achieving efficiency and high quality.

\begin{figure*}[htb]
    \centering
    \includegraphics[width=\linewidth]{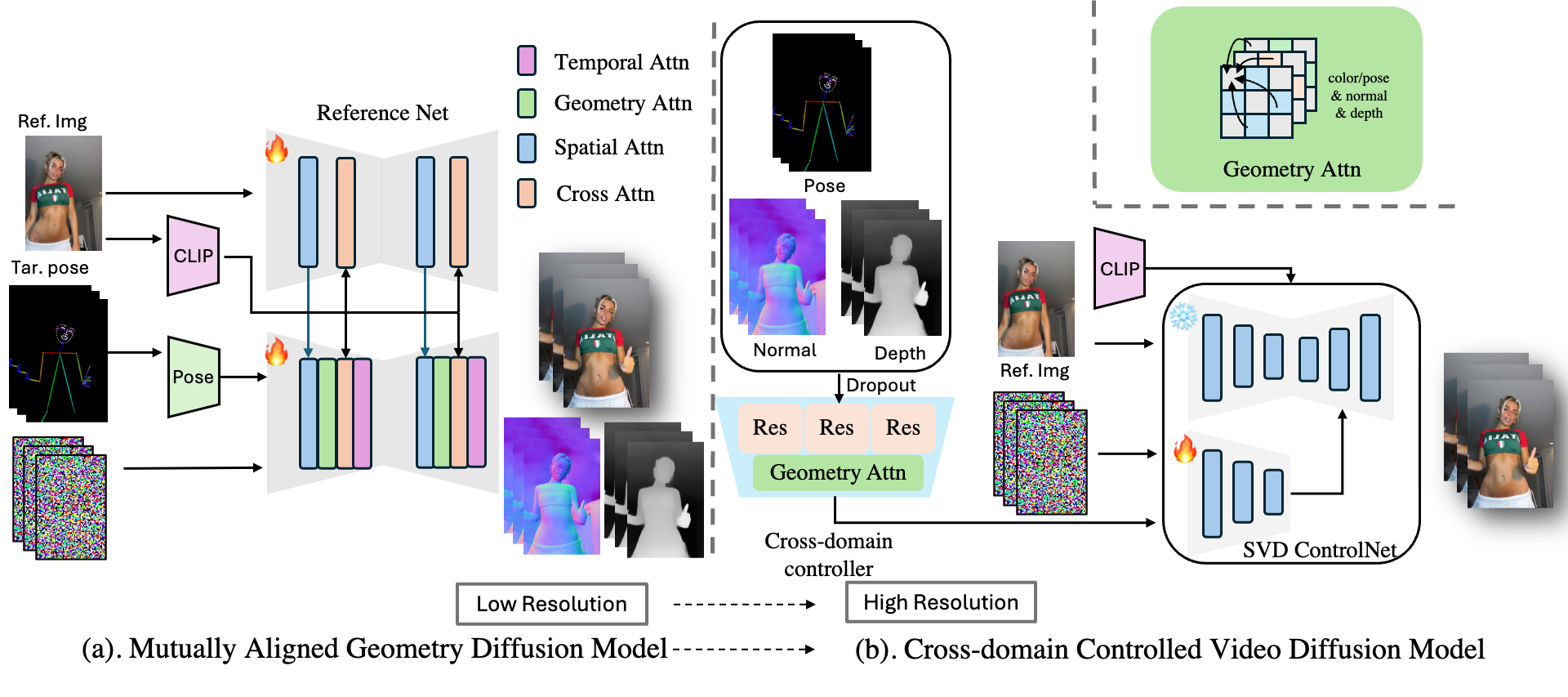}
    \vspace{-2.5em}

    \caption{
    \textbf{Overview of DreamDance framework.} The Mutually Aligned Geometry Diffusion Model generates detailed depth and normal maps to enrich guidance signals that are mutually aligned across modalities and time. The Cross-domain Controlled Video Diffusion Model utilizes a cross-domain controller to integrate multiple levels of guidance, producing high-quality human animations.}
\vspace{-10pt}
    \label{fig:frame}
\end{figure*}
\section{Preliminaries and Problem Formulation}
\subsection{Diffusion Models}
Diffusion models, as introduced in previous works~\cite{ho2020denoising,sohl2015deep}, aim to approximate a data distribution $q(\mathbf{x})$ by learning a probabilistic model $p_{\theta}(\mathbf{x}_{0}) = \int{p_{\theta}(\mathbf{x}_{0:T})\, \mathrm{d}\mathbf{x}_{1:T}}$. The joint distribution between $\mathbf{x}_{0}$ and random latent variables $\mathbf{x}_{1:T}$ is defined through a reverse Markov Chain, $p_\theta(\mathbf{x}_{0:T})=p(\mathbf{x}_T) \prod_{t=1}^T p_\theta(\mathbf{x}_{t-1}|\mathbf{x}_t)$, where $p(\mathbf{x}_T)$ is modeled as a standard normal distribution, and the transition kernels are expressed as $p_\theta(\mathbf{x}_{t-1}|\mathbf{x}_t)=\mathcal{N}(\mathbf{x}_{t-1};\mathbf{\mu}_\theta(\mathbf{x}_t,t),\sigma^2_t \mathbf{I})$. A forward Markov Chain is also constructed as $q(\mathbf{x}_{1:T}|\mathbf{x}_0)=\prod_{t=1}^{T} q(\mathbf{x}_t|\mathbf{x}_{t-1})$, allowing $\mathbf{\mu}_\theta(\mathbf{x}_t,t)$ to be defined as:
$\frac{1}{\sqrt{\alpha}_t}\left(\mathbf{x}_t - \frac{\beta_t}{\sqrt{1-\bar{\alpha}_t}} \mathbf{\epsilon}_\theta (\mathbf{x}_t, t)\right)$,
where the noise predictor $\mathbf{\epsilon}_\theta$ is trained using the loss function:

\begin{small}
\begin{equation}
\setlength\abovedisplayskip{-5pt}
\ell=\mathbb{E}_{t,\mathbf{x}_0,\mathbf{\epsilon}}\left[\|\mathbf{\epsilon} - \mathbf{\epsilon}_\theta (\sqrt{\bar{\alpha}_t} \mathbf{x}_0+\sqrt{1-\bar{\alpha}_t}\mathbf{\epsilon}, t)\|_2\right],
\end{equation}
\end{small}

\noindent with $\bar{\alpha}_t$ as a constant and $\mathbf{\epsilon}$ sampled from a standard normal distribution.
Building on the foundation of diffusion models, latent diffusion models~\cite{rombach2022high} extend this approach by mapping data into a low-dimensional latent space using a pre-trained variational auto-encoder (VAE)~\cite{esser2021taming}. The diffusion process is then applied within this latent space, which significantly reduces computational costs and allows these models to handle larger datasets and more complex tasks, such as video and 3D generation. In our work, we fine-tune our diffusion models using these advanced latent diffusion models and their variants to achieve our specific goals.

\subsection{Problem Formulation}
\label{sec:pro}

Given a human image $X$ and a sequence of driving poses $p_{1:T}$, we aim to generate a human animation video $Y_{1:T}$. We propose to capture the correlations in a coarse-to-fine manner, facilitating both intra-frame coherence and inter-frame consistency. Specifically, we first jointly generate fine-grained image $x_t$, depth $d_t$, and normal maps $n_t$ in low-resolution space to enrich guidance for each frame $t$ with a Mutually Aligned Geometry Diffusion Model $G_1$. Then we incorporate multiple guidance to animate the human image utilizing a high-resolution Cross-domain Controlled Video Diffusion Model $G_2$ with a cross-domain controller. Note that we drop the low-resolution RGB images $x_{1:T}$ generated in the first stage. The overall pipeline can be formulated as,

\begin{small}
\begin{equation}
\setlength\abovedisplayskip{-5pt}
\setlength\belowdisplayskip{-5pt}
\begin{aligned}
        &{x_{1:T}, n_{1:T}, d_{1:T}} = G_1^{low\_res}(X, p_{1:T}) \\
        &Y_{1:T} = G_2^{high\_res}(X, p_{1:T}, n_{1:T}, d_{1:T})
\end{aligned}   
\end{equation}
\end{small}

\section{Method: DreamDance}
We propose DreamDance, a unified framework for animating human images with high quality and precise control. As shown in Figure~\ref{fig:frame}, our method consists of two stages according to our problem formulation in Section~\ref{sec:pro}. Given the reference image and target poses, the Mutually Aligned Geometry Diffusion Model generates the target normal and depth maps as well as RGB images in the first stage, enriching geometry cues as guidance. Subsequently, the Cross-Domain Controlled Video Diffusion Model utilizes a cross-domain controller to integrate multiple levels of guidance for high-quality human image animating. 
\subsection{Mutually Aligned Geometry Diffusion Model}
\label{sec:4.1}

To enrich geometric information from the reference image given the target poses, we aim to first establish correlations that transition from a coarse-level pose skeleton to fine-grained spatial geometry. However, several challenges arise. 1). The original diffusion model is trained on RGB images, which differ significantly from geometry representations, making it difficult to model these disparate modalities within a unified diffusion process. 2). While the target pose provides the basic human structure, it lacks detailed geometry and appearance information that should be sourced from the reference image. 3). Ground truth RGB images are inherently aligned with their geometric features, necessitating the preservation of this alignment throughout the generation process to ensure coherence and accuracy.

\noindent \textbf{Modeling a Unified Diffusion Process.} As shown in Figure~\ref{fig:frame} (a)., we model a unified diffusion process that jointly generates depth $\mathbf{d}$ and normal maps $\mathbf{n}$, along with RGB images $\mathbf{x}$, all aligned with the target poses $\mathbf{p}$ based on the reference images $\mathbf{i}$. The unified diffusion model can be trained with a simplified objective:

% \begin{equation}
% \begin{aligned}
% \ell &=\mathbb{E}_{t,\mathbf{x,n,d,i,p},\mathbf{\epsilon}} \left[ \|\mathbf{\epsilon} - \mathbf{\epsilon}_\theta (\mathbf{x_t;i,p})\|_2^2 \right. \\  &\left.  + \|\mathbf{\epsilon} - \mathbf{\epsilon}_\theta (\mathbf{n_t;i,p})\|_2^2 + \|\mathbf{\epsilon} - \mathbf{\epsilon}_\theta (\mathbf{d_t;i,p})\|_2^2 \right],
% \end{aligned}
% \end{equation}

\vspace{-10pt}
\begin{small}
\begin{equation}
\begin{aligned}
& \ell =\mathbb{E}_{t,\mathbf{z,i,p},\mathbf{\epsilon}} \left[ \|\mathbf{\epsilon} - \mathbf{\epsilon}_\theta (\mathbf{z_t;i,p})\|_2^2 \right. \\
& \mathbf{z_t} = concat(\mathbf{x_t},\mathbf{n_t},\mathbf{d_t})
\end{aligned}
\label{eq:uni}
\end{equation}
\end{small}
% \vspace{-15pt}

\noindent where $\epsilon \sim \mathcal{N}(0,I)$ are independent Gaussian noise and $t$ is the sampled timestep that determines the scale of added noise. The noisy latents are obtained by $\mathbf{x_t/n_t/d_t}=\alpha_t(\mathbf{x/n/d}) + \sigma_t\epsilon$ and are concatenated along the batch dimension as a unified noisy latent $\mathbf{z_t}$, which is sent into the denoising UNet to predict the added noise. Given that the original diffusion model has a strong prior in the image domain, which differs from the distribution of depth and normal maps, we introduce a domain embedding to the UNet architecture. This aims to simplify the training process by better accommodating the distinct characteristics of each domain. Specifically, we use a one-hot vector to specify the domain of each sample, which is then encoded using positional encoding. The resulting domain embedding is added to the time embedding in the UNet. We find that incorporating the domain embedding leads to more stable training and faster convergence. 

\noindent \textbf{Control with Reference Image and Target Pose.} Inspired by \cite{hu2024animate}, we introduce a reference UNet for detailed reference image feature guidance. The spatial attention in the denoising UNet is modified as, 

\vspace{-10pt}
\begin{small}
\begin{equation}
\begin{aligned}
    \mathbf{q} &= W_q \cdot \mathbf{x}, \\
    \mathbf{k} &= W_k \cdot concat(\mathbf{x, x_{ref}}), \\
    \mathbf{v} &= W_v \cdot concat(\mathbf{x, x_{ref}}),
\end{aligned}
\end{equation}
\end{small}
% \vspace{-10pt}

\noindent where $\mathbf{x}$ is the noisy features and $\mathbf{x_{ref}}$ is the detailed injection features from the reference UNet. These features are concatenated along the sequence dimension.  Additionally, to guide the diffusion model with the target pose, we employ a lightweight convolutional pose encoder with zero convolution projection to encode the pose embedding following~\cite{hu2024animate,zhu2024champ}. The pose embedding is added to the noisy latent features after the conv\_in layer of the UNet.

\noindent \textbf{Mutual Alignment with Geometry Attention.} 
We now extend the original diffusion model to jointly generate normal and depth maps, along with RGB images, all semantically aligned with the target pose and reference image. However, there is no inherent guarantee that these generated outputs will be mutually consistent. To address this, we propose a geometry attention module that aligns the geometric representations and RGB images during the diffusion process, ensuring coherence across the generated content. The geometry attention is modified based on the self-attention mechanism in which the queries, keys, and values for each different modality are computed as follows,

\vspace{-10pt}
\begin{small}
\begin{equation}
\begin{aligned}
    \mathbf{q_i} &= W_q \cdot \mathbf{x_i},\quad \mathbf{q_n} = W_q \cdot \mathbf{x_n}, \quad \mathbf{q_d} = W_q \cdot \mathbf{x_d}, \\
    \mathbf{k_i} &= W_k \cdot cat(\mathbf{x_i, x_n, x_d}), \mathbf{v_i} = W_v \cdot cat(\mathbf{x_i, x_n, x_d}), \\
    \mathbf{k_n} &= W_k \cdot cat(\mathbf{x_i, x_n, x_d}), \mathbf{v_n} = W_v \cdot cat(\mathbf{x_i, x_n, x_d}), \\
    \mathbf{k_d} &= W_k \cdot cat(\mathbf{x_i, x_n, x_d}), \mathbf{v_d} = W_v \cdot cat(\mathbf{x_i, x_n, x_d}), \\
\end{aligned}
\label{eq}
\end{equation}
\end{small}
% \vspace{-10pt}

\noindent where $cat()$ refers to the concatenation operation along the sequence dimension.
The geometry attention module enhances mutual-guided geometric consistency, ensuring the generated geometric information remains stable and reliable. Additionally, we incorporate a temporal attention module to smooth the generated content, promoting temporal consistency. By aligning geometric information across both modality and temporal dimensions, we establish a robust foundation for effective control in the subsequent stages of video generation.

\noindent \textbf{Training Strategy.}  The training process is conducted in three steps. First, we disable the geometry and temporal attention to train the model using random inputs from different modalities. This allows the model to learn independently from each modality without the complexity of cross-modal and temporal interactions. Next, the geometry attention is activated with the other modules freeze, focusing on aligning the different modalities. Finally, we activate the temporal attention and freeze the other modules, ensuring temporal consistency. This three-step strategy stabilizes the training process by gradually introducing the complexity of multi-modal alignment and temporal consistency, ensuring stable convergence.
At this stage, all the training steps are conducted at a relatively low resolution to balance efficiency and performance effectively.  We empirically find this strategy can not only stabilize the process and accelerate convergence but also result in minimal performance loss compared to training at high resolution.  

\noindent \textbf{Multi-domain CFG.} Classifier-free guidance~\cite{ho2022classifier} (CFG) is a technique that enables a balance between sample quality and diversity for diffusion models. Our empirical observations indicate that different domains generated by our diffusion model require distinct optimal guidance scales for best results, especially for normal maps. Thus we propose multi-domain CFG(MCFG) to apply different guidance scales to each domain based on Equation~\ref{eq:uni} and formulate as follows,

\begin{small}
\begin{equation}
\setlength\abovedisplayskip{-10pt}
\begin{aligned}
    &\mathbf{\epsilon}_\theta(\mathbf{x_t;i,p}) ,\mathbf{\epsilon}_\theta(\mathbf{n_t;i,p}), \mathbf{\epsilon}_\theta(\mathbf{d_t;i,p}) = chunk( \mathbf{\epsilon}_\theta(\mathbf{z_t;i,p})) \\
    &\Tilde{\mathbf{\epsilon}}_\theta(\mathbf{x_t;i,p}) = \mathbf{\epsilon}_\theta(\mathbf{x_t;i,p}) + s_x\cdot(\mathbf{\epsilon}_\theta(\mathbf{x_t;i,p})-\mathbf{\epsilon}_\theta(\mathbf{x_t;\varnothing,p})) \\
    &\Tilde{\mathbf{\epsilon}}_\theta(\mathbf{n_t;i,p}) = \mathbf{\epsilon}_\theta(\mathbf{n_t;i,p}) + s_n\cdot(\mathbf{\epsilon}_\theta(\mathbf{n_t;i,p})-\mathbf{\epsilon}_\theta(\mathbf{n_t;\varnothing,p})) \\
    &\Tilde{\mathbf{\epsilon}}_\theta(\mathbf{d_t;i,p}) = \mathbf{\epsilon}_\theta(\mathbf{d_t;i,p}) + s_d\cdot(\mathbf{\epsilon}_\theta(\mathbf{d_t;i,p})-\mathbf{\epsilon}_\theta(\mathbf{d_t;\varnothing,p}))
\end{aligned}
\end{equation}
\end{small}
\vspace{-10pt}

% One of the reasons is that video data lacks ground-truth depth and normal maps, requiring pre-processing with pre-trained estimators to generate pseudo ground-truth. However, the estimation process often introduces more artifacts at higher resolutions. Meanwhile,

\subsection{Cross-domain Controlled Video Diffusion Model}
With the enriched guidance signals (i.e., depth and normal maps) mutually consistent and aligned with the target pose, we now aim to integrate these various types of guidance into a video diffusion model to animate human images. Note that we drop the low-resolution RGB images generated in the previous stage.

\noindent \textbf{Cross-domain Controller}
As shown in Figure~\ref{fig:frame} (b)., We introduce a cross-domain controller to integrate multiple guidance signals, enabling high-fidelity and accurate animation of human images. Specifically, each control signal modality—such as depth, normal maps, and pose, is first embedded into the feature space through a series of lightweight, domain-specific convolutional layers. This step captures the essential features of each modality. Next, we employ a similar geometry attention mechanism, as outlined in Equation~\ref{eq}, to integrate these feature vectors, ensuring harmonious interaction across modalities. This unified guidance feature drives the animation process with coherence and precision. Formally, the guidance feature $\mathbf{f}$ for each frame is obtained by 

\vspace{-10pt}
\begin{small}
\begin{equation}
    \mathbf{f_i} = GeoAttn(F_p(\mathbf{p_i}), F_d(\mathbf{d_i}), F_n(\mathbf{n_i})) , i\in [0:T]
\end{equation}
\end{small}

\vspace{-10pt}

\noindent where $F_p, F_d, F_n$ are the domain-specific convolution encoders. The guidance feature is then added to the noisy latent features after the conv\_in layer of the SVD ControlNet.

\begin{table*}[htbp!]
 \small
  \centering
    \begin{subtable}[t]{\linewidth}
    \centering
  \begin{tabular}{lccccccccc}
    \toprule
    \multirow{2}{*}{Method~~}&\multirow{2}{*}{Original}&\multicolumn{4}{c}{Image} &\multicolumn{2}{c}{Video} \\
    \cmidrule(r){3-6} \cmidrule(r){7-8}  & Guidance &L1\(\downarrow\)~~ & PSNR\(\uparrow\)~~  &SSIM\(\uparrow\)~~ &LPIPS\(\downarrow\)~~  & FID-VID\(\downarrow\)~~& FVD\(\downarrow\)~~\\ 
    \midrule
    MRAA~\cite{siarohin2021mraa} &2D &3.21E-04 & 29.39& 0.672& 0.296&54.47 &284.82 \\
    DisCo~\cite{wang2023disco} &2D & 3.78E-04 & 29.03& 0.668& 0.292& 59.90 & 292.80 \\
    AnimateAnyone~\cite{hu2024animate} &2D & - & 29.56& 0.718& 0.285&- &171.90\\
    MagicAnimate~\cite{xu2024magicanimate} &2D & 3.13E-04 & 29.16& 0.714& 0.239&\underline{21.75} &179.07 \\
    Champ~\cite{zhu2024champ} &2D+3D &\underline{3.02E-04} & \underline{29.84}& \underline{0.773}& \underline{0.235}& 26.14 &\underline{170.20} \\
    \textbf{Ours} & 2D & \textbf{2.89E-04} & \textbf{29.90} & \textbf{0.798}& \textbf{0.233}& \textbf{19.86} &\textbf{153.07}\\
    \bottomrule
  \end{tabular}
  \caption{Quantitative comparisons on TikTok dataset.}
  \label{tab:comp:tiktok}
  \end{subtable}
  % \vspace{\fill}
  \begin{subtable}[t]{\linewidth}
  \centering
  \begin{tabular}{lccccccc}
    \toprule
    \multirow{2}{*}{Method~~}&\multirow{2}{*}{Original}&\multicolumn{4}{c}{Image} &\multicolumn{2}{c}{Video} \\
    \cmidrule(r){3-6} \cmidrule(r){7-8}  & Guidance &L1\(\downarrow\)~~ & PSNR\(\uparrow\)~~  &SSIM\(\uparrow\)~~ &LPIPS\(\downarrow\)~~  & FID-VID\(\downarrow\)~~& FVD\(\downarrow\)~~\\ 
    \midrule
    AnimateAnyone~\cite{hu2024animate} &2D &3.25E-04 & 29.37& 0.725& 0.278&\underline{22.69} &\underline{177.38} \\
    MagicAnimate~\cite{xu2024magicanimate} &2D &3.47E-04 & 29.09& 0.712& 0.291& 28.23 &196.77 \\
    Champ~\cite{zhu2024champ} &2D+3D &\underline{3.19E-04} & \underline{29.60}& \underline{0.746}& \underline{0.243}&23.83 &182.25\\
    Ours &2D & \textbf{2.83E-04 }&\textbf{29.85}& \textbf{0.788}& \textbf{0.233}& \textbf{20.37} & \textbf{158.24}\\

    \bottomrule
  \end{tabular}
  \caption{Quantitative comparisons on the proposed dataset.}
  \label{tab:comp:pp}
  \end{subtable}
  \vspace{-0.9em}
  \caption{Quantitative comparisons with baselines, with best results in \textbf{bold} and second best results \underline{underlined}.}
  \label{tab:comp}
\vspace{-1.2em}
\end{table*}

\noindent \textbf{SVD ControlNet.} To effectively control the video diffusion model generation process, we adapt the ControlNet~\cite{zhang2023adding} on the SVD model. Specifically, SVD ControlNet freezes all parameters of the pre-trained SVD, while keeping a trainable copy of selected layers from the original network. These two branches are linked by zero-initialized convolution layers that gradually incorporate controllable features during training. Empirically, we find that using SVD ControlNet leads to more stable training compared to fine-tuning the entire network.

\noindent \textbf{Robust Conditioning.}
Since the generated depth and normal conditions from the first stage may contain artifacts, error accumulation becomes a potential issue in our two-stage pipeline, potentially leading to performance degradation. To mitigate this, we propose a simple yet effective dropout strategy for the control signals to enhance robust conditioning. Specifically, we randomly replace control signals across temporal dimensions and modalities with zero-value images, encouraging the model to leverage cues from alternative modalities and temporal frames rather than strictly adhering to the conditions. This approach significantly improves robustness during inference and proves especially effective in enhancing temporal consistency.

\section{Experiments}

\subsection{TikTok-Dance5K Dataset}
Large-scale datasets with high-quality samples are important for video generation tasks. We construct a human dance dataset consisting of around 5K videos to facilitate high-fidelity human image animation. All the samples are collected from TikTok and followed by a manual cleaning process. We preprocess the dataset to obtain pseudo ground truth pose, normal, and depth maps~\cite{yang2023effective,depthanything,bae2024dsine}.  The dataset will be made publicly available.

\begin{figure}[h]
    \centering
    \includegraphics[width= \linewidth]{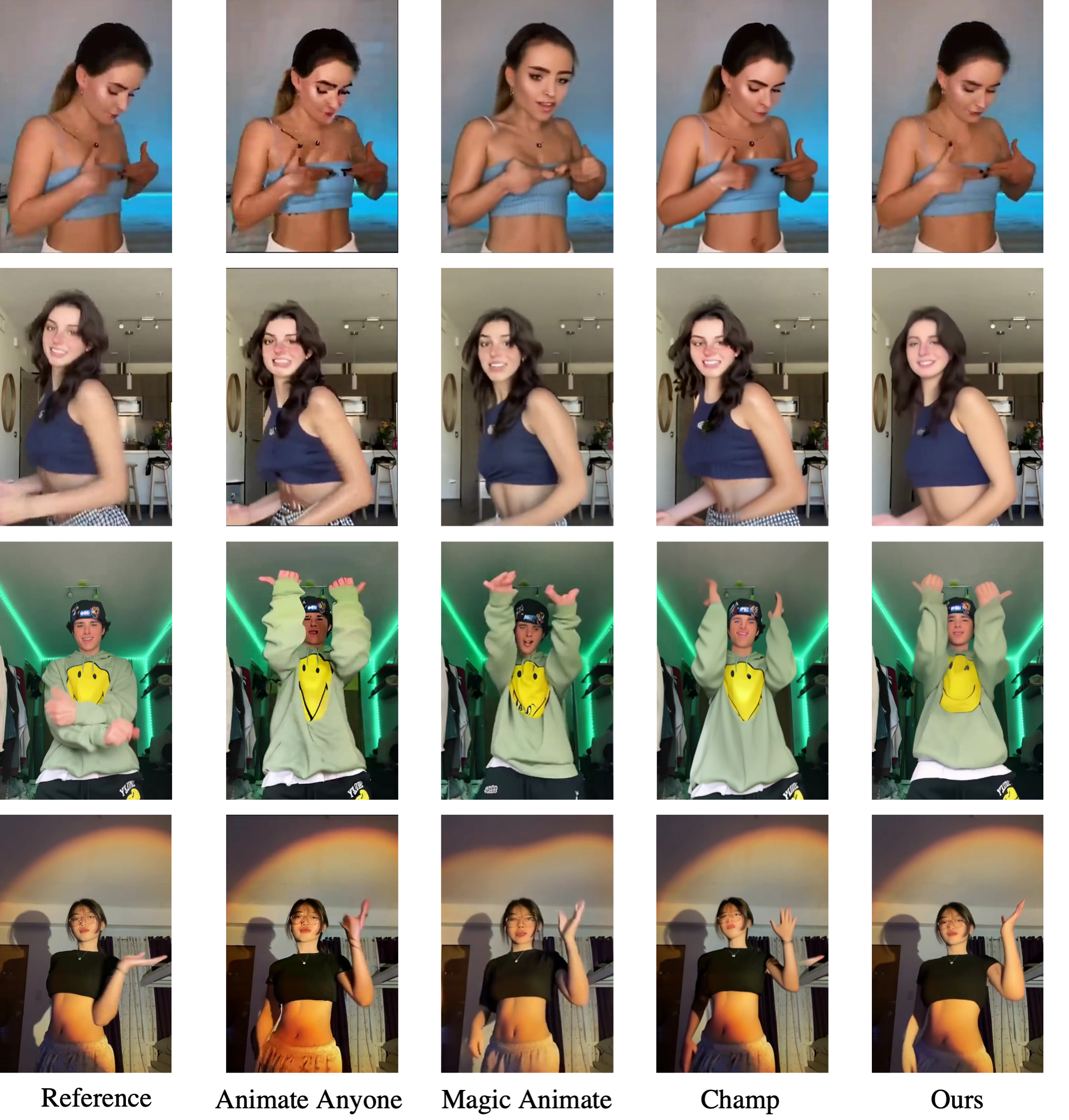}
    \vspace{-2em}
    \caption{Qualitative results comparing to baseline methods. The first two rows are from the TikTok dataset and the last two are from the proposed dataset's testing set.}
    \label{fig:compare}
    \vspace{-2em}
\end{figure}

\subsection{Implementation Details}
Our experiments are conducted on 8 NVIDIA A800 GPUs. To train the geometry diffusion model in the first stage, we initialize both the reference UNet and denoising UNet with Stable Diffusion v1.5. The training resolution is 256x384. As for the video diffusion model in the second stage, we initialize the model with SVD v1.1 and train the model for 50,000 steps with a batch size of 8. The training video is cropped and resized to the resolution of 512x768 and comprises 16 frames. For more details on the training settings, please refer to the Appendix.

\subsection{Main Results}
\noindent \textbf{Baselines.} MRAA~\cite{siarohin2021mraa} is the state-of-the-art GAN-based animation method that warps the source image by estimating the optical flow of the driving sequence and then inpaints the occluded regions using a GAN model. DisCo~\cite{wang2023disco} is a baseline diffusion-based animation method that integrates a module of separation conditions for pose, body, and background into a pre-trained diffusion model for human image animation. MagicAnimate~\cite{xu2024magicanimate} and AnimateAnyone~\cite{hu2024animate} are diffusion-based human image animating methods that employ 2D control signals as guidance. Champ~\cite{zhu2024champ} is developed based on AnimateAnyone and develops a 3D representation such as SMPL to model the control sequence. The SMPL representation is rendered into depth and normal maps as well as semantic maps. Then combined with 2D skeleton poses, multiple control signals are fused and together guide the animating process. 

\noindent \textbf{Evaluation Metrics.}
Following the evaluation method in previous works, we adopt L1 error, Structural Similarity Index (SSIM)~\cite{wang2004ssim}, Learned Perceptual Image Patch Similarity (LPIPS)~\cite{zhang2018lpips}, and Peak
Signal-to-Noise Ratio (PSNR)~\cite{hore2010psnr} to evaluate single-frame image quality. To evaluate video fidelity, we use Frechet Inception Distance with Fréchet Video Distance (FID-FVD)~\cite{balaji2019fidvid} and Fréchet Video Distance (FVD)~\cite{unterthiner2018fvd}.

\noindent \textbf{Evaluation on the Benchmark Dataset.} We evaluate our proposed method on the benchmark dataset TikTok dataset~\cite{Jafarian_2021_CVPR_TikTok} and report metrics in Table~\ref{tab:comp:tiktok}. Our method shows state-of-the-art performance compared to baseline methods, achieving lower L1 losses, LPIPS, FID-VID, FVD scores, and higher PSNR, and SSIM values. We also provide qualitative results in Figure~\ref{fig:compare}. 

\begin{figure}[ht]
    \centering
    \includegraphics[width= \linewidth]{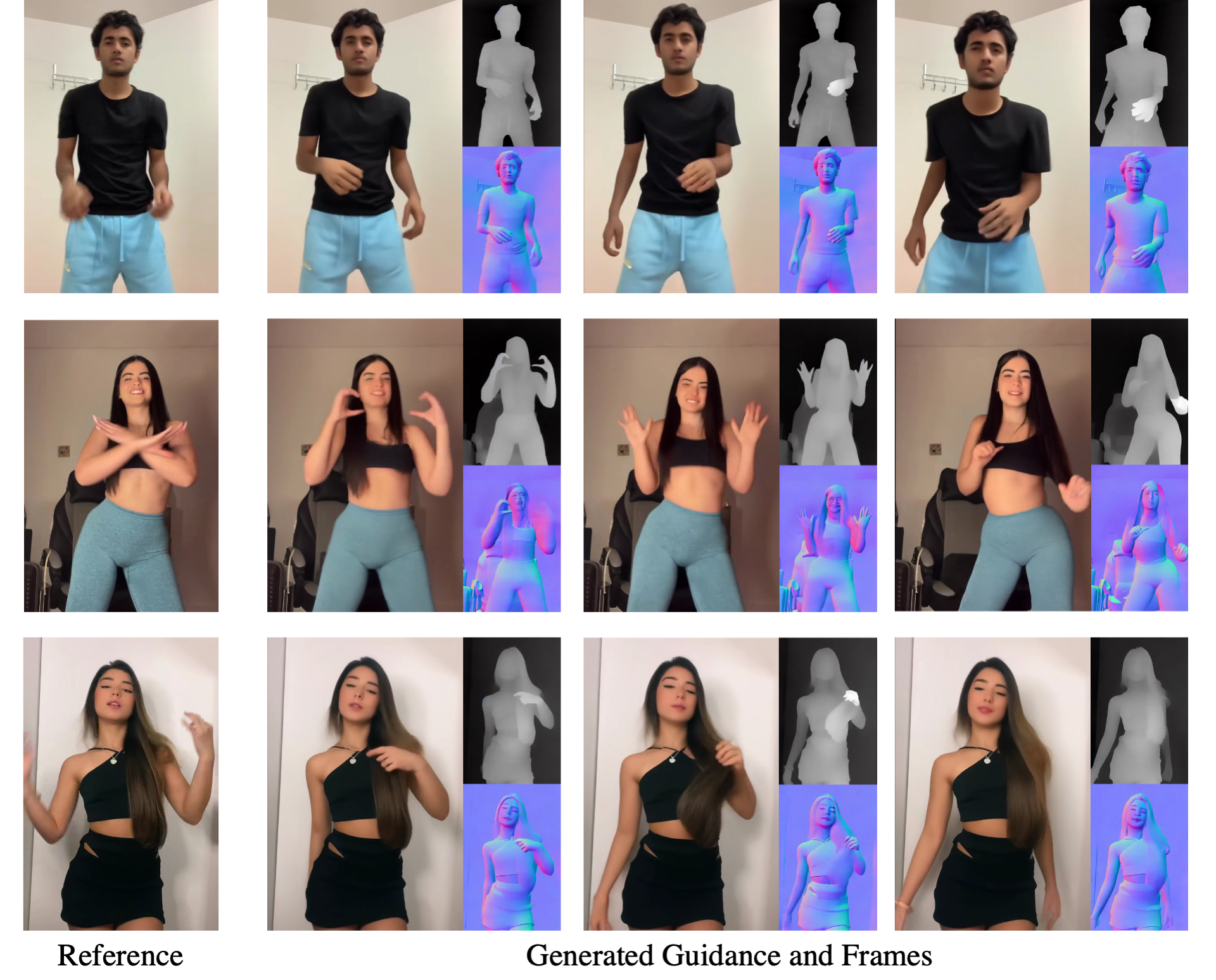}
    \vspace{-2em}
    \caption{Qualitative results with generated normal and depth.}
    \label{fig:more}
    \vspace{-0.8em}
\end{figure}

\noindent \textbf{Evaluation on the Proposed Dataset.}
To further evaluate our proposed method, we conduct experiments on the testing set of our proposed dataset. We report the evaluation metrics in Table~\ref{tab:comp:pp} and present qualitative comparisons in Figure~\ref{fig:compare}. Our method achieves state-of-the-art performance compared to various baseline methods. We present more qualitative results with generated normal and depth guidance in Figure~\ref{fig:more}. We observe that both AnimateAnyone~\cite{hu2024animate} and MagicAnimate~\cite{xu2024magicanimate} produce low-quality results due to insufficient guidance, while Champ~\cite{zhu2024champ} fail to generate accurate details, especially in the hand area. This is because SMPL struggles to reconstruct complex hand poses, leading to imprecise guidance map renderings. 

\begin{figure}[ht]
    \centering
    \includegraphics[width= \linewidth]{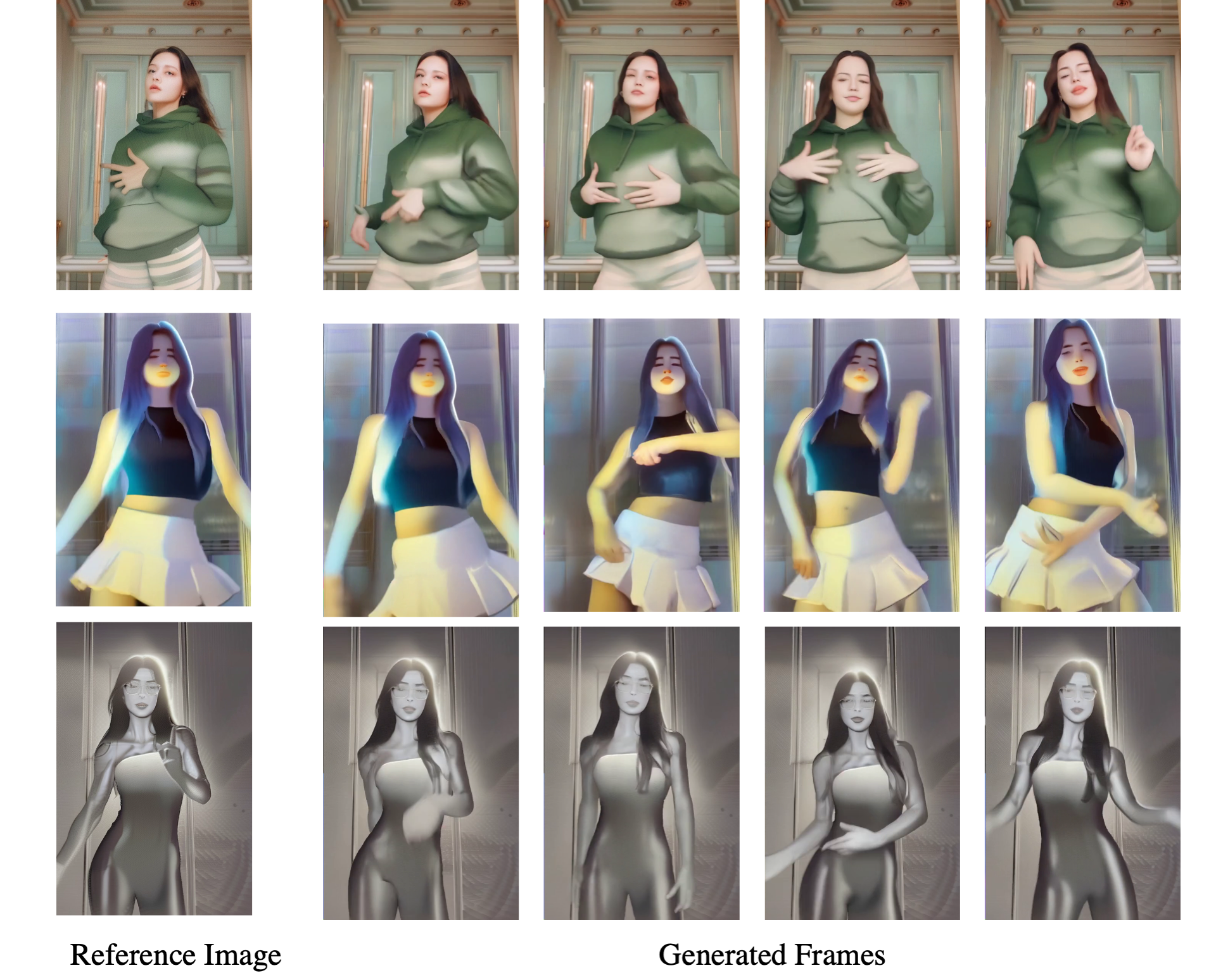}
    \vspace{-2em}
    \caption{Qualitative results of animating unseen domain images.}
    \vspace{-1em}
    \label{fig:unseen}
\end{figure}

\noindent \textbf{Animating Unseen Domain Images.} We present qualitative results of animating images from unseen domains in Figure~\ref{fig:unseen}, using the obtained normal and depth conditions. These results demonstrate the generalization capabilities of the proposed method.

\begin{figure}[h]
    \centering
    \includegraphics[width= \linewidth]{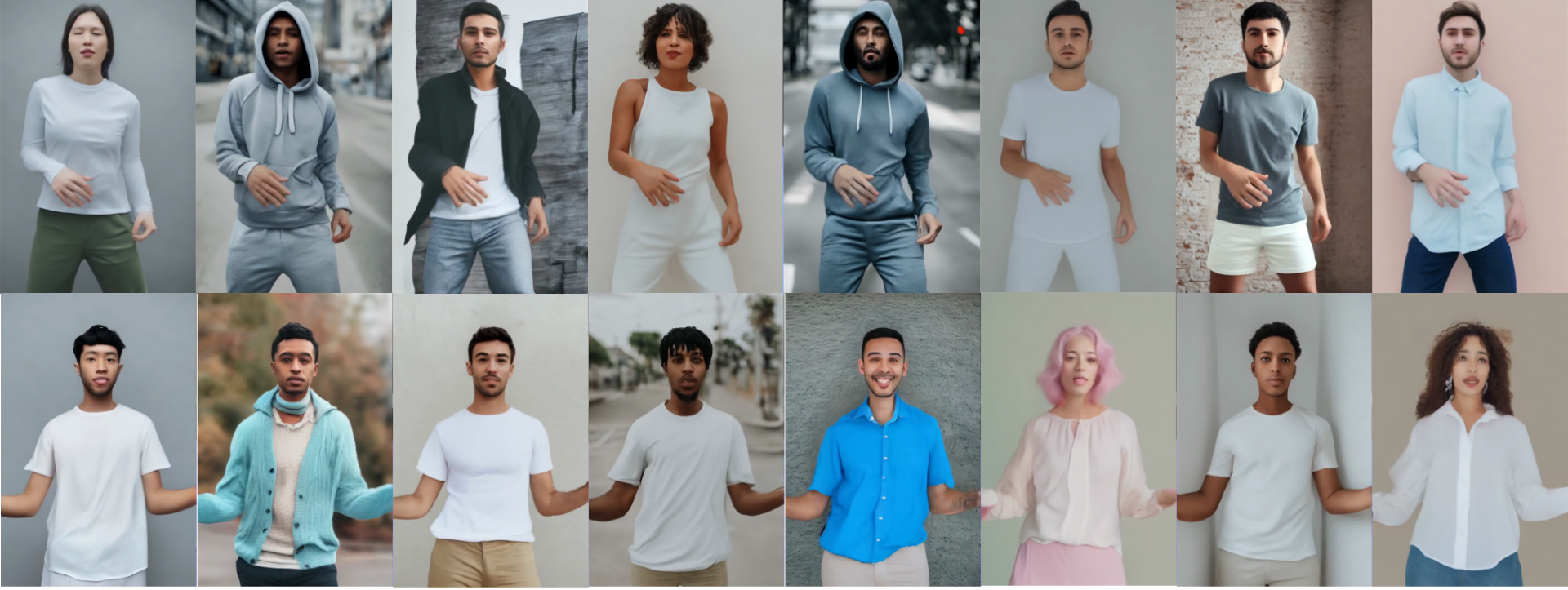}
    \vspace{-1.5em}
    \caption{Qualitative results of cross-ID animation.}
    \label{fig:xid}
    \vspace{-1.5em}
\end{figure}

\noindent \textbf{Cross-ID Animation.} We animate different human images with the same pose sequence and present in Figuire~\ref{fig:xid}.

\begin{table}[b]
\vspace{-1.5em}
\small
    \centering
    \begin{tabular}{c|ccc}
    \toprule
        Method & Stage 1 &  & Total\\
    \midrule
         Ours&  1.13s & & 1.13s\\
    \bottomrule
    \toprule
        Method & 3D prediction & shape transfer & Total \\        
        & \&smooth & \&rendering & \\
        \midrule
        Champ~\cite{zhu2024champ}& 0.55s & 0.43s & 0.98s \\ 
    \bottomrule
    \end{tabular}
    % \vspace{-1em}
    \caption{Inference efficiency analysis for obtaining geometry guidance. We report average inference time per frame in seconds. }
    \label{tab:eff}
    \vspace{-0.5em}
\end{table}

\begin{table*}[t]
    \small
    \centering
    \begin{subtable}[t]{\linewidth}
    \centering
    \begin{tabular}{cccccccc}
    \toprule
    Guidance &L1\(\downarrow\)~~ & PSNR\(\uparrow\)~~  &SSIM\(\uparrow\)~~ &LPIPS\(\downarrow\)~~  & FID-VID\(\downarrow\)~~& FVD\(\downarrow\)~~ \\
    \midrule
    a). Without Pose & 3.38E-04 & 29.43 & 0.743& 0.255& 22.38 &175.37\\
    b). Without Depth& 3.95E-04 & 29.02 & 0.701& 0.277& 24.56 &193.23 \\
    c). Without Normal &3.67E-04 & 29.24 & 0.723& 0.263& 23.28 &183.84 \\
    d). Pose+Normal+Depth & 2.89E-04 & 29.90 & 0.798& 0.233& 19.86 &153.07  \\

    \bottomrule
    \end{tabular}
    \caption{Ablation studies on different conditions.}
    \label{tab:ablation_map}
    \end{subtable}
        \begin{subtable}[t]{\linewidth}
    \centering
    \begin{tabular}{cccccccc}
    \toprule
    Controller &L1\(\downarrow\)~~ & PSNR\(\uparrow\)~~  &SSIM\(\uparrow\)~~ &LPIPS\(\downarrow\)~~  & FID-VID\(\downarrow\)~~& FVD\(\downarrow\)~~ \\
    \midrule
    Without GeoAttn & 3.36E-04 & 29.48 & 0.767& 0.242& 21.93 & 165.27\\

    With GeoAttn &2.89E-04 & 29.90 & 0.798& 0.233& 19.86 &153.07 \\

    \bottomrule
    \end{tabular}
    \caption{Ablation studies on the effectiveness of geometry attention.}
    \label{tab:ablation_geo}
    \end{subtable}
    \vspace{-1em}
    \caption{Quantitative ablation studies.}
    \vspace{-2em}
  \label{tab:abl}
\end{table*}

\noindent \textbf{Inference Efficiency Analysis.}
Table~\ref{tab:eff} presents the inference efficiency analysis of our proposed method. We report the average time consumption of obtaining all the guidance maps when only 2D pose skeletons are given. Results show that our method consumes a similar time compared to Champ~\cite{zhu2024champ}. Considering Champ requires multiple steps when obtaining guidance maps, including 3D prediction, 3D smoothness, shape transfer, and rendering, our method that generates all the guidance maps using a single diffusion model is more efficient and user-friendly.

\subsection{Ablation studies}

\begin{figure}[h]
    \centering
    \includegraphics[width= \linewidth]{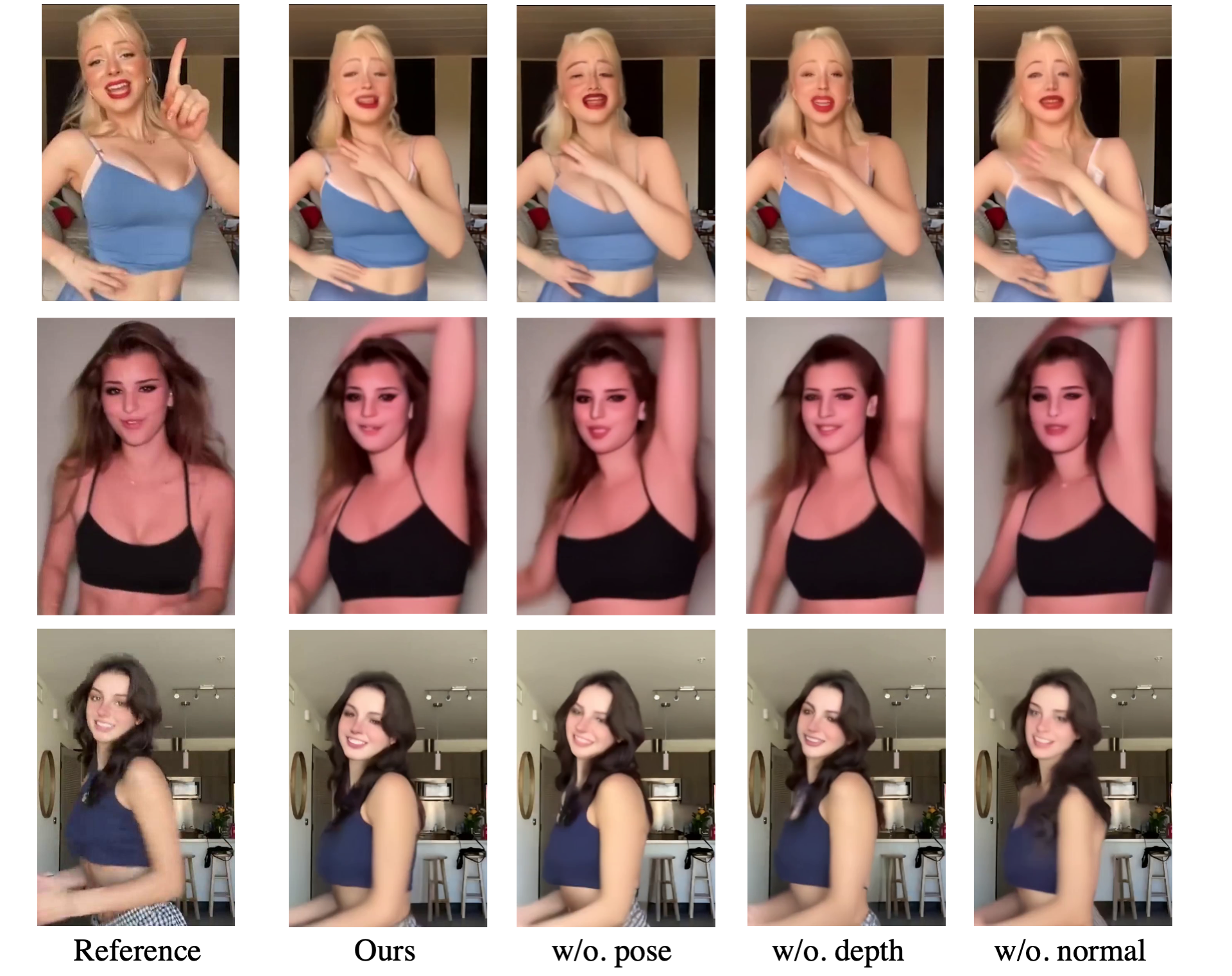}
    \vspace{-2em}
    \caption{Qualitative ablation studies on different conditions.}
    \vspace{-1.5em}
    \label{fig:abl}
\end{figure}

\noindent \textbf{Different Conditions.} We conduct comprehensive experiments on different variants of the conditions incorporated in the video diffusion model to show the effectiveness of the proposed method. As shown in Table~\ref{tab:ablation_map}, comparing different variants a), b), c) with our method d), we show that the incorporation of multiple conditions achieves robust and best performance. Qualitative comparisons are presented in Figure~\ref{fig:abl}. We observe that without depth or normal maps, the quality of the generation is noticeably flawed. Meanwhile, as originally provided, the pose maps provide refined guidance for facial and hand regions, improving generation quality.

\begin{figure}[h]
    \centering
    \includegraphics[width= \linewidth]{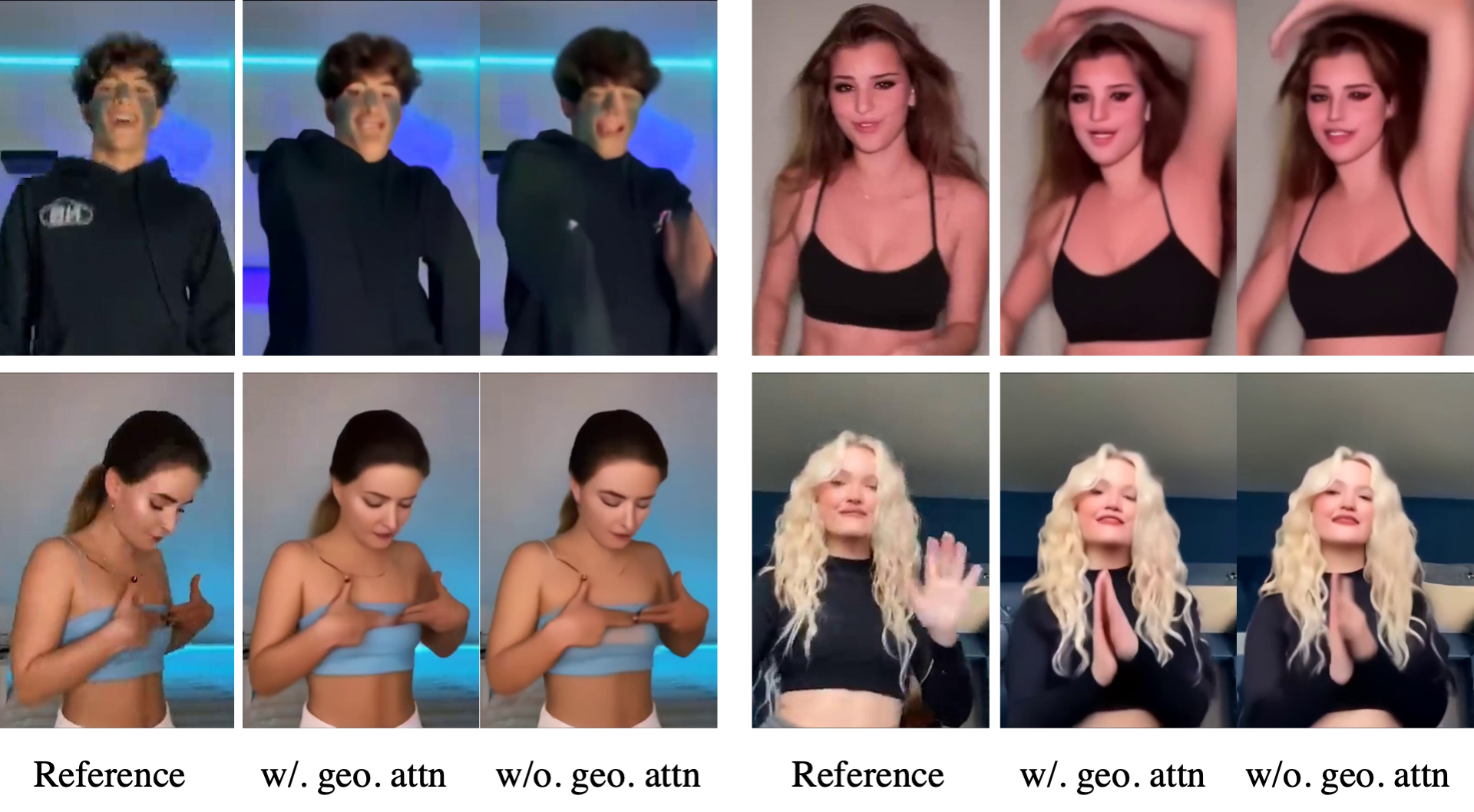}
    \vspace{-2em}
    \caption{Qualitative results of ablation studies on the effectiveness of geometry attention.}
    \label{fig:geo}
    \vspace{-1em}
\end{figure}

\begin{figure}[h]
    \centering
    \includegraphics[width= \linewidth]{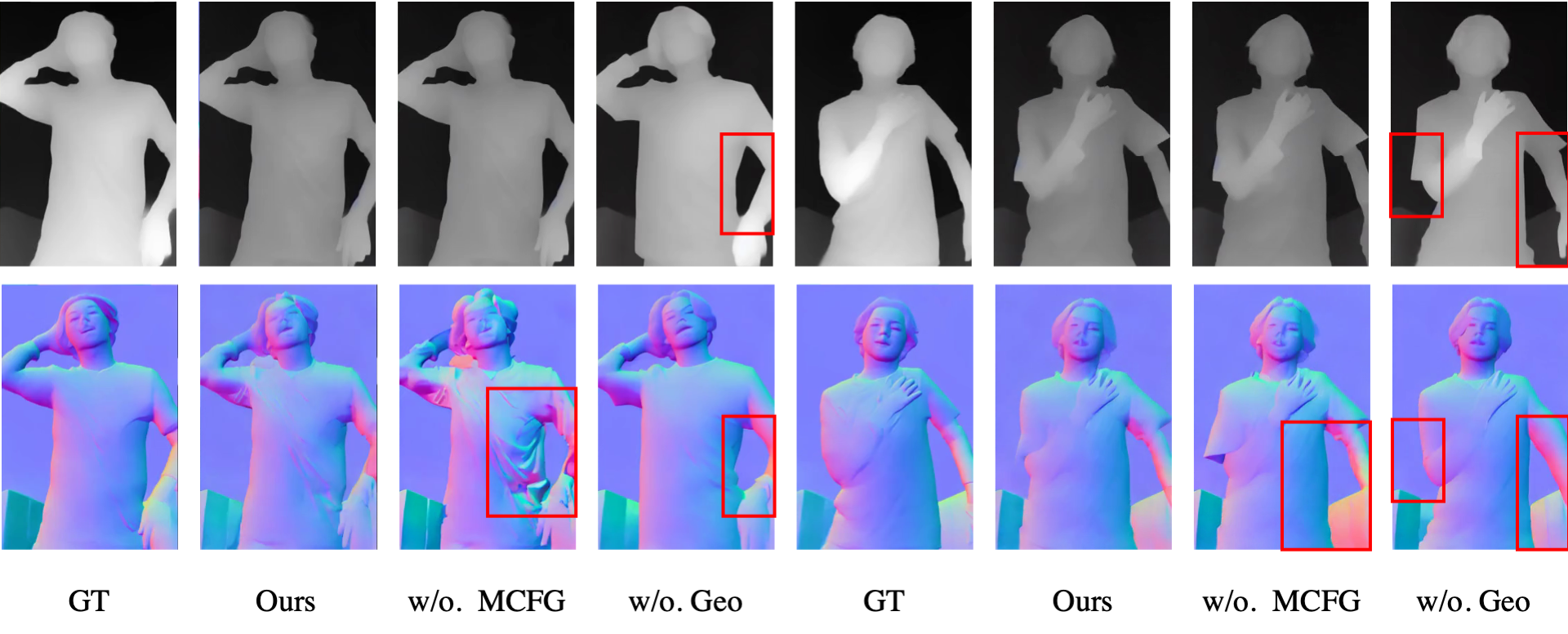}
    \vspace{-2em}
    \caption{Qualitative ablation studies on the effectiveness of proposed modules for geometry alignment. (Please zoom in to view.)}
    \label{fig:ab_geo}
    \vspace{-1em}
\end{figure}

\noindent \textbf{Geometry Alignment.}
To validate the effectiveness of the proposed geometry attention and the MCFG strategy for geometry alignment in the first stage, we show qualitative ablation studies in Figure~\ref{fig:ab_geo}. Without MCFG, the generated normal maps show an over-saturated color that cannot represent normal information effectively. Without the geometry attention, normal and depth maps are not aligned with each other.

\noindent \textbf{Geometry Attention for Conditions Fusion.}
We conduct an ablation study to evaluate the effectiveness of the proposed geometry attention mechanism for conditions fusion. Quantitative comparisons are presented in Table~\ref{tab:ablation_geo}, demonstrating that including geometry attention significantly enhances performance. Additionally, qualitative comparisons in Figure~\ref{fig:geo} further illustrate its benefits. We observe geometry attention effectively integrates multiple types of conditions, leading to more robust and high-quality generation.

\section{Conclusion}
This paper introduces DreamDance, a unified framework for animating human images using skeleton pose sequences. By employing innovative two-stage diffusion models, including the Mutually Aligned Geometry Diffusion Model to enrich guidance signals and the Cross-domain Controlled Video Diffusion Model to integrate multiple levels of guidance, DreamDance significantly enhances visual coherence and temporal consistency of the generated video. Extensive experiments demonstrate that the proposed method achieves state-of-the-art performance, offering a flexible and user-friendly solution for realistic human image animating.

% \clearpage

% \clearpage
{
    \small
    \bibliographystyle{ieeenat_fullname}
    \bibliography{main}
}


\end{document}